\documentclass[letterpaper]{article} 
\usepackage{aaai25}  
\usepackage{times}  
\usepackage{helvet}  
\usepackage{courier}  
\usepackage[hyphens]{url}  
\usepackage{graphicx} 
\usepackage{multirow}
\usepackage{natbib}  
\usepackage{caption} 
\usepackage{algorithm}
\usepackage{algpseudocode}

\usepackage{amsmath}
\usepackage{amssymb}

\usepackage{newfloat}
\usepackage{listings}
\DeclareCaptionStyle{ruled}{labelfont=normalfont,labelsep=colon,strut=off} 
\floatstyle{ruled}
\newfloat{listing}{tb}{lst}{}
\floatname{listing}{Listing}
\title{Motion-aware Contrastive Learning for \\ Temporal Panoptic Scene Graph Generation}
\author{
    Thong Thanh Nguyen\textsuperscript{\rm 1}, \;
    Xiaobao Wu\textsuperscript{\rm 2}, \;
    Yi Bin\textsuperscript{\rm 1,3\thanks{Yi Bin is the corresponding author, yi.bin@hotmail.com}}, \; \\ 
    Cong-Duy Nguyen\textsuperscript{\rm 2}, \;
    See-Kiong Ng\textsuperscript{\rm 1}, \;
    Anh Tuan Luu\textsuperscript{\rm 2}
}
\affiliations{
    \textsuperscript{\rm 1} Institute of Data Science (IDS), National University of Singapore, Singapore \\
    \textsuperscript{\rm 2} Nanyang Technological University (NTU), Singapore, \textsuperscript{\rm 3} Tongji University, China\\

}

\usepackage{bibentry}
\usepackage{color}
\definecolor{orange}{RGB}{253,107,9}

\newcommand{\answerTODO}[1][]{\textcolor{red}{\bf [TODO]}}

\begin{document}

\maketitle

\begin{abstract}
To equip artificial intelligence with a comprehensive understanding towards a temporal world, video and 4D panoptic scene graph generation abstracts visual data into nodes to represent entities and edges to capture temporal relations. Existing methods encode entity masks tracked across temporal dimensions (mask tubes), then predict their relations with temporal pooling operation, which does not fully utilize the motion indicative of the entities' relation. To overcome this limitation, we introduce a contrastive representation learning framework that focuses on motion pattern for temporal scene graph generation. Firstly, our framework encourages the model to learn close representations for mask tubes of similar subject-relation-object triplets. Secondly, we seek to push apart mask tubes from their temporally shuffled versions. Moreover, we also learn distant representations for mask tubes belonging to the same video but different triplets. Extensive experiments show that our motion-aware contrastive framework significantly improves state-of-the-art methods on both video and 4D datasets. \end{abstract}

%

\section{Introduction}

The advent of autonomous agents, intelligent systems, and robots warrants a comprehensive understanding of real-world environments \citep{ma2022sqa3d, driess2023palm, raychaudhuri2023reduce, cheng2022masked, li2023tube, li2023transformer}. Such understanding encompasses beyond merely recognizing individual entities, but also a sophisticated understanding of their relationships. To construct a detailed understanding, scene graph generation (SGG) research \citep{li2022dynamic, sudhakaran2023vision, nag2023unbiased, wang2024oed} has sought to provide relational perspective on scene understanding. In SGG frameworks, scene graphs utilize nodes to represent entities and edges to represent relationships, constructing a comprehensive and structured understanding of visual scenes. 

However, due to being primarily based on bounding boxes to denote entities, scene graphs fall short of replicating human visual perception with a lack of granularity \citep{yang2023panoptic}. To overcome this limitation, panoptic scene graph generation \citep{yang2022panoptic, zhao2023textpsg} has been presented to expand the scope of SGG to incorporate pixel-level precise entity localization and thorough scene understanding including background components. Subsequently, because the temporal dimension undoubtedly provides richer information than the static spatial dimension, recent works \citep{yang2023panoptic, yang20244d} have shifted attention to the domain of videos and 4D scenes, resulting in the tasks of panoptic video and 4D scene graph generation.

\begin{figure}[t]
\centering
\includegraphics[width=\linewidth]{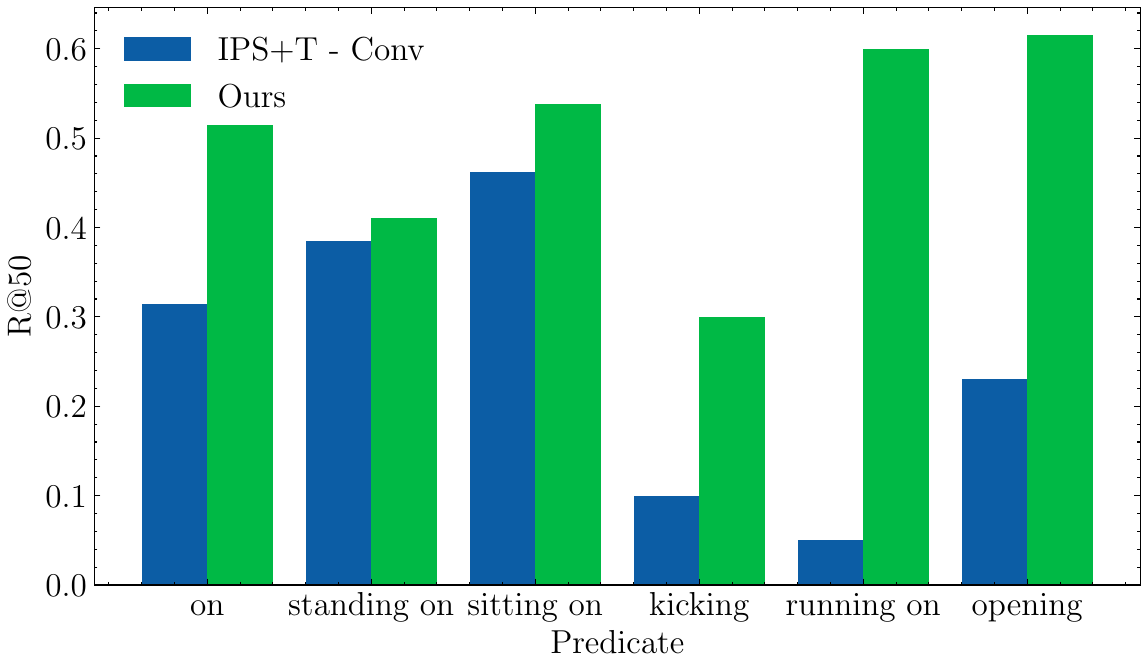} 
\caption{State-of-the-art IPS+T - Convolution \citep{yang2023panoptic} exhibits high R@50 scores for static relations, \textit{e.g.} \textit{on}, \textit{sitting on}, and \textit{standing on}, than dynamic relations, \textit{e.g.} \textit{kicking}, \textit{running on}, and \textit{opening}. In contrast, our method can perform effectively on both static and dynamic relations.}
\label{fig:r@50}
\vspace{-15pt}
\end{figure}

\begin{figure*}[t]
\centering
\includegraphics[width=\linewidth]{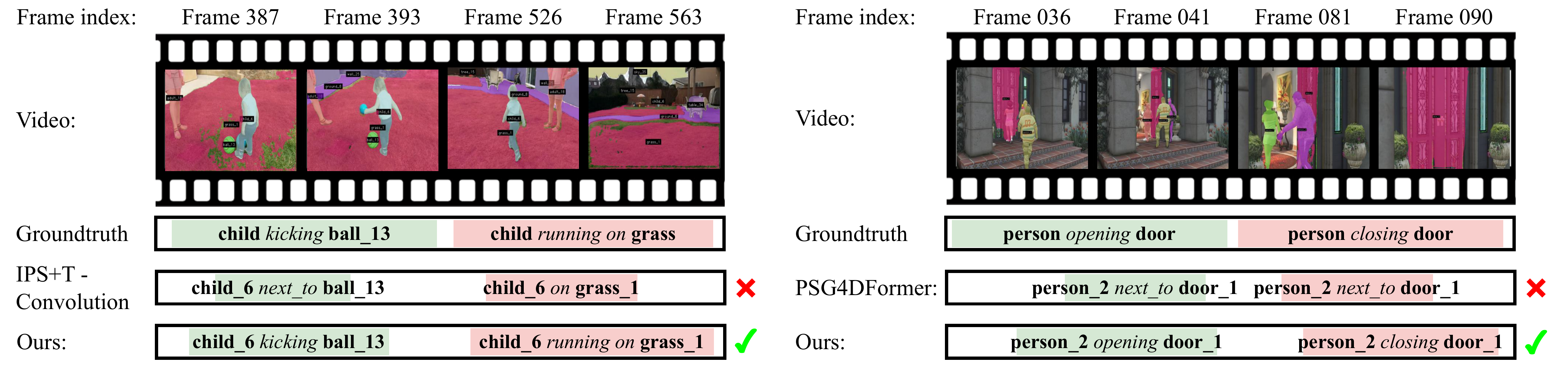} 
\caption{Examples of temporal panoptic scene graph generation of state-of-the-art  \citep{yang2023panoptic, yang20244d} and our method.}
\label{fig:qualitative_example}
\vspace{-15pt}
\end{figure*}

Popular methods \citep{yang2023panoptic, yang20244d, yang2022panoptic} for temporal panoptic scene graph generation produce entity masks tracked across the temporal dimension, \textit{i.e.} mask tubes, then predict temporal relations among them. To conduct relation prediction, these methods encode the segmentation mask tubes, apply global pooling, then forward to a multi-layer perceptron for classifying their relations. However, such global pooling operation is well-known to be ineffective in representing temporal or motion patterns, which are useful for determining the relation among the entities. Consequently, this would result in higher misclassification rates of more dynamic relations \citep{wang2023taking, nag2023unbiased, zhou2022context}, as illustrated in Figure \ref{fig:r@50}. 

To encourage temporal representation learning, current research \citep{nguyen2023demaformer, liu2022spatial, zhou2023learning} uses contrastive learning for videos. However, they mainly seek to force two clips from the same video to be close together. As such, they mostly capture the semantics of visual scenes and disregard motions \citep{chen2020graph}. Moreover, instead of paying attention to precise entity localization, they work upon frame-level representations. This would inadvertently inject motions from non-target entities into visual representations, which might not benefit relation classification in panoptic scene graph generation.

In this paper, to encourage representation learning to capture motion patterns for temporal panoptic scene graph generation, we propose a novel contrastive learning framework that focuses on mask tubes of the segmented entities. First, we force a mask tube and the one of similar subject-relation-object but of a different video to obtain close representations. Since positive mask tubes originate from distinct video clips, the model cannot rely upon visual semantics to optimize the contrastive objective, but instead depends on the motion trajectory evolution, which is our target component for representation learning. Second, we propel negative mask tubes generated by temporally shuffling the original tubes. Moreover, we also push apart representations of mask tubes from the same video but belonging to different triplets. Because mask tubes of different triplets from a common video with close visual features are separated from each other, we once again motivate the model to generate representations that are less reliant upon visual semantics but motion-sensitive features. In addition, the visually similar negative mask tubes can play a role as hard negative samples, thus accelerating the contrastive learning process \citep{chen2024curriculum}.

Moreover, in order to implement our motion-aware contrastive learning framework, there is a need to quantify the relationship between mask tubes. This quantification marks a challenging problem as mask tubes are a sequence of segmentation masks that span over the sequence of video frames. Furthermore, mask tubes of two triplets might exhibit different lengths since two events often occur at different speed. Unfortunately, the popular pipeline of temporal pooling and then similarity estimation flattens the temporal dimension of the mask tubes and neglects their motion features. To resolve this problem, we consider mask tubes of two triplets as two distributions and seek the optimal transportation map between them, then utilize the transport distance as the distance between two triplets' tubes. Such scheme of transporting can play a role of synchronizing the motion states of two triplets and takes advantage of the mask tubes' evolutionary trajectory. 

To sum up, our contributions are as follows:
\begin{itemize}
    \item We propose a novel contrastive learning framework for temporal panoptic scene graph generation which pulls together entity mask tubes with similar motion patterns and pushes away those of distinct motion patterns.
    \item We utilize optimal transport distance to estimate the relationship between two events' mask tubes for the proposed contrastive framework.
    \item Comprehensive experiments demonstrate that our framework outperforms state-of-the-art methods on both natural and 4D video datasets, especially on recognizing dynamic subject-object relations.
\end{itemize}
\section{Related Work}
\noindent\textbf{Temporal panoptic scene graph generation.} Traditional research mainly focuses on generating scene graphs for natural videos in which nodes represent objects and edges represent relations between objects. To localize object instances, the nodes are grounded by bounding boxes \citep{wang2024oed, nag2023unbiased, pu2023spatial}. Despite progress, limitations exist in traditional video scene graph generation because of noisy grounding annotations of coarse bounding box annotations and trivial relation taxonomy. Recent works have addressed these issues by proposing panoptic video and 4D scene graph generation \citep{yang2023panoptic, yang20244d}. Classic video scene graph generation methods have been dominated by the two-stage pipeline that consists of object detection and pairwise predicate classification \citep{rodin2024action, nag2023unbiased}.
This property has been generalized to panoptic approaches \citep{yang2022panoptic, yang2023panoptic, yang20244d}, in which the pipeline comprises panoptic segmentation followed by predicate classification step.

\noindent\textbf{Video representation learning.} Video representation learning has gained popularity in recent years \citep{nguyen2024read, nguyen2024encoding}. Most approaches can be categorized into two groups: pretext-based and contrastive-based. Pretext-based methods mostly leverage pretext learning tasks such as optical flow prediction \citep{dong2024memflow, davtyan2023efficient} and temporal order prediction \citep{shen2024patch, ren2024arvideo}. However, these tasks are considerably influenced by low-level features and incapable of delving into high-level semantics of the video \citep{nguyen2024video}. 
The contrastive-based methods primarily construct positive samples \citep{nguyen2021contrastive, wu2023infoctm, nguyen2024kdmcse, nguyen2022adaptive, wu2024modeling,nguyen2024topic, nguyen2023improving} by sampling video clips from the same video \citep{nguyen2024meta, liu2024contrastive}, use various frame-based data augmentation techniques \citep{wang2024havtr, song2024animation, rosa2024video}, etc., thereby increasing the similarity of positive pairs while simultaneously decreasing the one of negative pairs. Nevertheless, these methods focus on the whole video frames and are not suitable for object masks tracked across the temporal dimension.
\vspace{-5pt}
\section{Problem Formulation}
Temporal panoptic scene graph generation (TPSGG) is a task to generate a dynamic scene graph given an input video. In the generated scene graph, each node corresponds to an entity and each edge corresponds to a spatial-temporal relation between two entities. Formally, the input of a TPSGG model is a video clip $V$, particularly $V \in \mathbb{R}^{T \times H \times W \times 3}$ for a natural video, $V \in \mathbb{R}^{T \times H \times W \times 4}$ for a 4D RGB-D video, and $V \in \mathbb{R}^{T \times M \times 6}$ for a 4D point cloud video, $T$ denotes the number of frames, $M$ the number of point clouds of interest, and the frame size $H \times W$ should remain consistent across the video. The output of the model is a dynamic scene graph $G$. The TPSGG task can be formulated as follows:
\begin{equation}
P(G|V) = P(M, O, R|V).
\end{equation}
In particular, G consists of binary mask tubes $M = \{\mathbf{m}_{1}, \mathbf{m}_{2}, ..., \mathbf{m}_{N}\}$ and entity labels $O = \{o_{1}, o_{2}, ..., o_{N}\}$ which are associated with $N$ entities in the video, and their relations are denoted as $R = \{r_{1}, r_{2}, ..., r_{L}\}$. With respect to entity $o_{i}$, the mask tube $\mathbf{m}_{i} \in \{0, 1\}^{T \times H \times W}$ composes all tracked masks in all video frames, and its category $o_{i} \in \mathbb{C}^{O}$. For all entities in frame $t$, their masks must not overlap, \textit{i.e.} $\sum\limits_{i=1}^{N} \mathbf{m}_{i}^{t} \leq \textbf{1}^{H \times W}$. The relation $r_{i} \in \mathbb{C}^{R}$ associates two entities, one of which is the subject and the other is an object, with a relation class and a time period. $\mathbb{C}^{O}$ and $\mathbb{C}^{R}$ denote the entity and relation set class, respectively.

\begin{figure*}[t]
\centering
\includegraphics[width=0.8\linewidth]{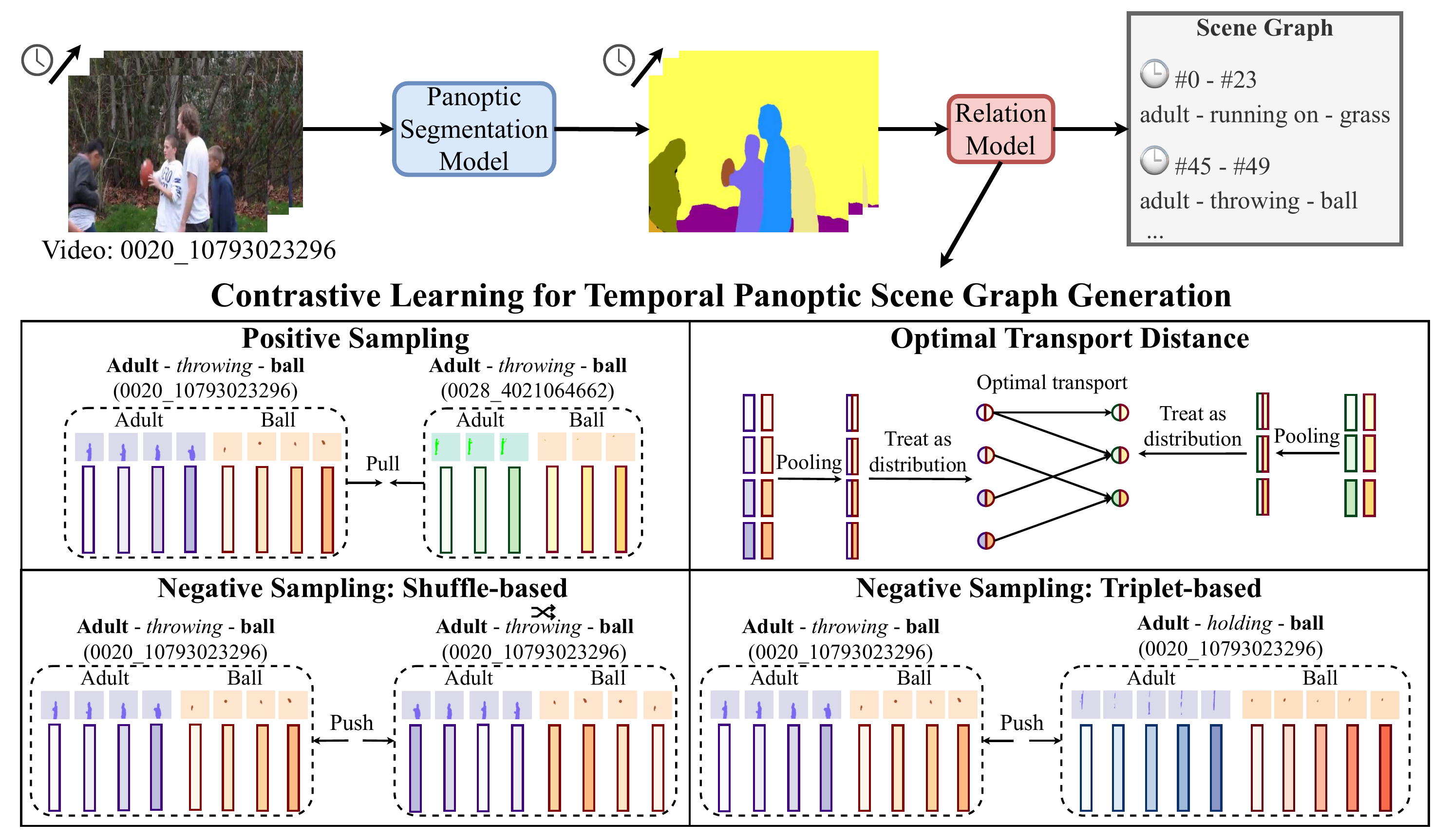} 
\caption{Framework overview of contrastive learning for temporal scene graph generation.}
\label{fig:framework}
\vspace{-10pt}
\end{figure*}
\vspace{-5pt}
\section{Methodology}
\begin{figure}[t]
\centering
\includegraphics[width=0.8\linewidth]{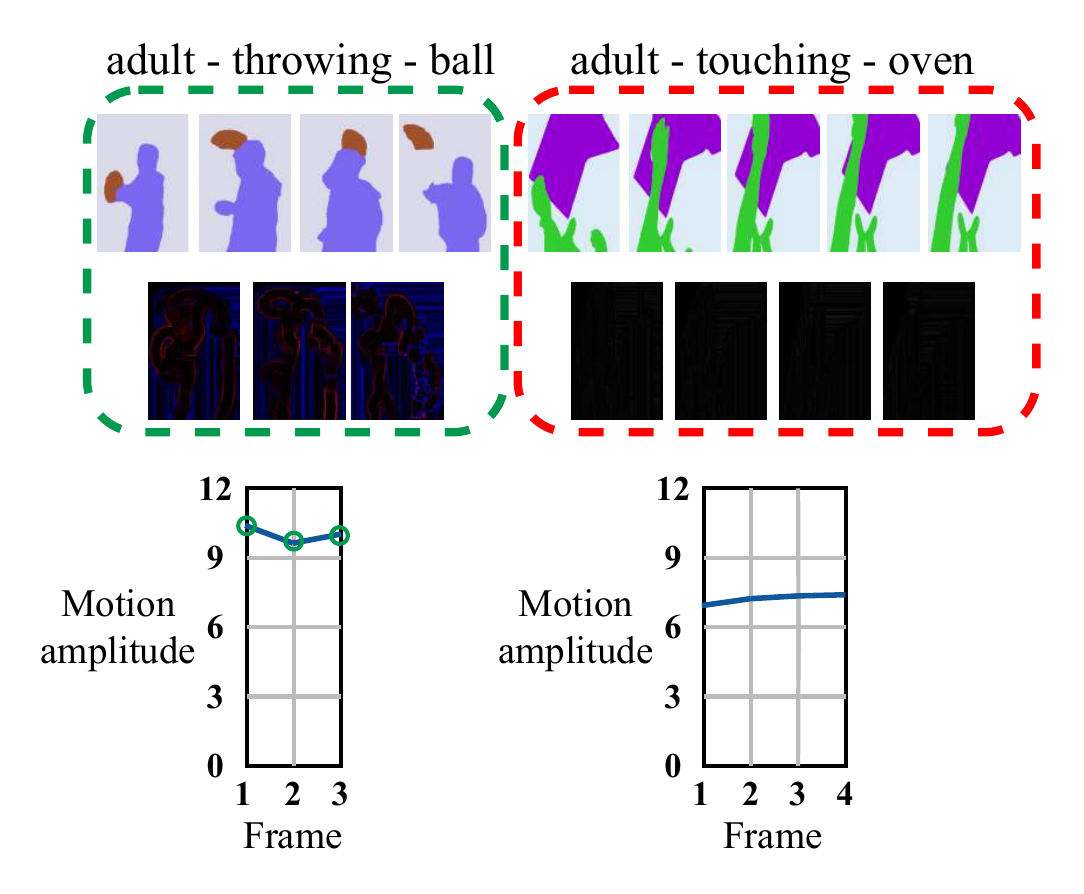} 
\caption{Proposed strategy to select strong-motion tubes.}
\label{fig:selecting_strong_motion}
\vspace{-15pt}
\end{figure}

We firstly present the backbone pipeline to conduct temporal panoptic scene graph generation. Then, we explain our proposed contrastive learning framework to facilitate motion-aware mask tube representation learning. We also present the detail of our optimal transport approach to estimate the relation between two mask tubes for the contrastive objective. Our overall framework is illustrated in Figure \ref{fig:framework}.

\subsection{Temporal Panoptic Segmentation}
Given a video clip $V \in \mathbb{R}^{T \times H \times W \times 3}$, $V \in \mathbb{R}^{T \times H \times W \times 4}$, or $V \in \mathbb{R}^{T \times M \times 6}$, the initial step is to segment and track each pixel in a non-overlapping manner. Formally, the model produces a set of entity masks $\{y_{i}\}_{i=1}^{N} = \{(\mathbf{m}_{i}, p_{i}(c))\}_{i=1}^{N}$, where $\mathbf{m}_{i} \in \{0,1\}^{T \times H \times W}$ denotes the tracked video mask, \textit{i.e.} the mask tube, and $p_{i}(c)$ denotes the probability of assigning class $c$ to the tube $\mathbf{m}_{i}$. $N$ denotes the number of entities, which consist of both foreground (thing) and background (stuff) classes.

\noindent\textbf{Segmentation module.} Inspired by \citep{yang2023panoptic, yang20244d}, we adopt the Transformer-based encoder-decoder segmentation model. There are two types of segmentation procedure: 1) image panoptic segmentation combined with a tracker (IPS+T) and 2) video panoptic segmentation (VPS). IPS+T procedure will process each video frame separately and uses the tracker to connect the mask tubes across the video frames, while VPS processes each video frame with its reference frame from a nearby timestamp. 

Both procedures are initiated by producing a set of object queries which interacts with encoded visual patches via masked cross-attention. Receiving a video $V$, the model produces a set of queries $\{\mathbf{q}_{i}\}_{i=1}^{N}$, where each query $\mathbf{q}_{i}$ corresponds to one entity. Subsequently, every query is forwarded to two multi-layer perceptrons (MLPs) to project the queries into mask classification and mask regression outputs. 

\noindent\textbf{Training and inference.} During training, each query is matched to a groundtruth mask through mask-based bipartite matching to calculate the segmentation loss. During inference, IPS+T generates panoptic segmentation masks for each frame, and uses the tracker to achieve $N$ tracked mask tubes. In contrast, VPS employs two query embeddings of the target and reference frame, and performs query-wised similarity tracking to obtain $N$ tracked mask tubes. 

\subsection{Relation Classification} 
After the segmentation step, if the relation module is to be trained, we match query tubes with the annotated groundtruth masks based on the tube IoU values with the groundtruth. Otherwise, we directly forward mask tubes to self-attention or convolutional layers for encoding them into hidden representations $\{H_{i}\}_{i=1}^{N}$, $H_{i} \in \mathbb{R}^{T \times D}$, where $D$ denotes the hidden dimension. Then, we construct query pairs from every two query tubes' representations $H_{i}$ and $H_{j}$, $i, j \in \{1, 2, ..., N\}, \; i \neq j$. Inspired by \citep{yang2023panoptic, yang20244d}, in every pair, we perform global pooling over the temporal dimension for each mask tube representation:
\begin{equation}
\mathbf{h}_{i} = \text{Pooling}\left(H_{i}\right),
\end{equation}
where $\mathbf{h}_{i} \in \mathbb{R}^{D}$. 	Afterwards, we concatenate $\mathbf{h}_{i}$ and $\mathbf{h}_{j}$, and forward to a MLP to generate the relation category:
\begin{equation}
\log p(r_{i,j}) = \text{MLP}\left(\left[\mathbf{h}_{i}, \mathbf{h}_{j}\right]\right).
\end{equation}
To train the relation classification module, we use the cross-entropy loss calculated based on the predicted relation log-likelihood and the groundtruth. For inference, we extract the relation of the highest log-likelihood.

\subsection{Contrastive Learning for Temporal Panoptic Scene Graph Generation}

Our goal is to encourage mask tube representations $\{H_{i}\}_{i=1}^{N}$ to become motion-aware. In the beginning, we concatenate the representations of two mask tubes $H_{i}^{\text{sub}}$ and $H_{j}^{\text{obj}}$, which have been matched to a groundtruth subject-relation-object triplet, to form an anchor representation $H_{i,j}$ (anchor):
\begin{equation}
H^{a}_{i,j} = [H_{i}^{\text{sub}}, H_{j}^{\text{obj}}],
\end{equation}
where $H^{a}_{i,j} \in \mathbb{R}^{T \times 2D}$. Then, we propose a contrastive learning framework in which we motivate the model to associate mask tubes based upon the motion information. The objective of contrastive learning is to produce a representation space through attracting positive pairs, \textit{i.e.} $H^{a}$ and $H^{p}$ (positive), while pushing apart negative pairs, \textit{i.e.} $H^{a}$ and $H^{n}$ (negative). We accomplish this by optimizing the contrastive objective, which is formulated as follows:
\begin{equation}
\mathcal{L}_{\text{cont}} = -\text{log}\frac{e^{\text{sim}\left(H^{a}, H^{p}\right)}}{e^{\text{sim}\left(H^{a}, H^{p}\right)} + \sum\limits_{z=1}^{N_{n}} e^{\text{sim}\left(H^{a}, H^{n}_{z}\right)}},
\end{equation}
where sim denotes the similarity function defined upon a pair of mask tube representations. The formulation shows that what the model learn is largely dependent upon how positive and negative samples are generated. 

\noindent\textbf{Positive sampling.} To satisfy our motion-aware requirement for contrastive learning, we extract mask tube representations from the entities of the same subject and object category that exhibit a similar groundtruth relation from another video. Since two videos possess distinct visual features, the model must rely on the shared motion pattern of similar subject-relation-object triplets to associate the anchor and the positive sample. 

\noindent\textbf{Negative sampling.} For negative sampling, we design two strategies, which result in two contrastive approaches, \textit{i.e.} shuffle-based and triplet-based contrastive learning.

\subsection{Shuffle-based contrastive learning} In our shuffle-based approach, we create negative samples by utilizing a series of temporal permutations $\pi$ to the anchor tube, \textit{i.e.} shuffling:
\begin{equation}
H^{n} = \pi\left(H^{a}\right).
\end{equation}
As such, the contrastive objective will force the model to propel representations of the anchor tube, which is in the normal order, from the shuffled tube, which exhibits a distorted motion due to the shuffled order. This would make the learned representation sensitive to frame ordering, \textit{i.e.} motion-aware, as the anchor $H^{a}$ and the negative tube $H^{n}$ share visual semantics and can only be distinguished using motion information. 

\noindent\textbf{Selecting strong-motion mask tubes.} However, there exists a potential risk: for static relations such as \textit{on}, \textit{next to}, and \textit{in}, mask tubes might involve almost no motion. As a result, the shuffled tube would become identical to the anchor one and the model would not be able to differentiate them and learn reasonably. To address this problem, we propose a strategy to select strong-motion tubes for shuffling, which we illustrated in Figure \ref{fig:selecting_strong_motion}.

Given a video, our aim is to select mask tubes that carry strong motion for shuffling. To measure the motion of the mask tube, we utilize optical flow edges \citep{xiao2021modist}. We estimate flow edges via employing a Sobel filter \citep{sobel2022sobel} onto the flow magnitude map and take the median over the flow edge pixels of the entity masks. Then, we select mask tubes whose the maximum value across the optical flow surpasses a threshold $\gamma$.

\subsection{Triplet-based contrastive learning}
To take advantage of motion-aware signals from triplets of similar subject-relation-object category, we design a triplet-based approach to create negative samples. A naive approach would be to sample mask tubes of any distinct subject-relation-object triplet from the anchor sample. However, if we run into triplets with all distincgt subject, relation, and object categories, the negative pair would be trivial for the model to distinguish, resulting in less effective learning.

In order to create harder negative samples, we choose negative mask tubes from the same video with the anchor. We create a multi-nomial distribution, where triplets that share more subject, relation, or object categories with the anchor will be more likely to be drawn. Hence, our negative samples can hold close visual semantics with the anchor sample, and increase the likelihood that the model depends on motion semantics to push them apart. From contrastive learning perspective, these samples form hard negative samples to accelerate the learning process \citep{chen2024curriculum}.

\subsection{Optimal Transport for Mask Tube Relation Quantification}
There is one remaining problem, \textit{i.e.} how to define the similarity function $\text{sim}$ for two mask tubes' representations $H_{i}$ and $H_{j}$. In this work, we consider two mask tubes as two discrete distributions $\boldsymbol{\mu}$ and $\boldsymbol{\nu}$, whose $H_{i}$ and $H_{j}$ are their supports, respectively. Formally, $\boldsymbol{\mu} = \sum\limits_{k=1}^{T_{i}} \mathbf{a}_{k} \delta_{\mathbf{h}_{i,k}}$ and $\boldsymbol{\nu} = \sum\limits_{l=1}^{T_{j}} \mathbf{b}_{l} \delta_{\mathbf{h}_{j,l}}$, where $\delta_{\mathbf{h}_{i,k}}$ and $\delta_{\mathbf{h}_{j,l}}$ denote the Dirac functions centered upon $\mathbf{h}_{i,k}$ and $\mathbf{h}_{j,l}$, respectively. The weights of the supports are $\mathbf{a} = \frac{\mathbf{1}_{T_{i}}}{T_{i}}$ and $\mathbf{b} = \frac{1_{T_{j}}}{T_{j}}$.

After defining the distribution scheme, we propose the tube alignment optimization problem, which is to find the transport plan that achieves the minimum distance between $\boldsymbol{\mu}$ and $\boldsymbol{\nu}$ as follows:
\vspace{-5pt}
\begin{gather}
d_{\text{OT}} = \mathcal{D}_{\text{OT}}(\boldsymbol{\mu}, \boldsymbol{\nu}) = \min_{\mathbf{T} \in \Pi(\mathbf{a}, \mathbf{b})} \sum\limits_{k=1}^{T_{i}} \sum\limits_{l=1}^{T_{j}} \mathbf{T}_{i,j} \cdot c\left(\mathbf{h}_{i,k}, \mathbf{h}_{j,l}\right), \\
\begin{split}
\text{s.t} \quad \Pi(\mathbf{a}, \mathbf{b}) = \{\mathbf{T} \in \mathbb{R}^{T_i \times T_j}_{+} \mid \mathbf{T}\mathbf{1}_{T_i} \leq \mathbf{a},  \mathbf{T}^{\top}\mathbf{1}_{T_j} \leq \mathbf{b}, \\\mathbf{1}_{T_i}^{\top} \cdot \mathbf{T} \cdot \mathbf{1}_{T_j} = s, \quad 0 \leq s \leq \min(T_i, T_j)\},
\label{eq:pvla_formulation}
\end{split}
\vspace{-5pt}
\end{gather}
where $c$ denotes a pre-defined distance between two vectors. We implement the cost distance $c\left(\mathbf{h}_{i,k}, \mathbf{h}_{j,l}\right) = 1 - \frac{\mathbf{h}_{i,k} \cdot \mathbf{h}_{j,l}}{||\mathbf{h}_{i,k}||_{2} ||\mathbf{h}_{i,k}||_{2}}$ as the cosine distance. As the exact optimization over the transport plan $\mathbf{T}$ is intractable, we adopt the Sinkhorn-based algorithm to estimate $\mathbf{T}$. We delineate the algorithm to calculate the distance in Algorithm \ref{alg:ot_distance}. To turn the distance into similarity value, we take its negative value and add to a pre-defined margin $\alpha$:
\begin{equation}
    \text{sim}\left(\mathbf{h}^{a}, \mathbf{h}^{p}\right) = \alpha - d_{\text{OT}}.
\vspace{-10pt}
\end{equation}

\setlength{\textfloatsep}{5pt}
\begin{algorithm}[t]
{\footnotesize \caption{Computing the optimal transport distance}
\label{alg:ot_distance}
}
{\footnotesize
\begin{algorithmic}
\Require{$\mathbf{C} = \{\mathbf{C}_{l,k} = c\left(\mathbf{h}_{i,l}, \mathbf{h}_{j,k}\right) \mid 1 \leq i \leq N_V, 1 \leq j \leq N_L\} \in \mathbb{R}^{T_i \times T_j}$,\;  $\mathbf{a} \in \mathbb{R}^{T_i},\; \mathbf{b} \in \mathbb{R}^{T_j},\; s,\; N_{\text{iter}}$} \\
$d_{\text{OT}} = \infty$ 
\For{$s=1$ to $\min(T_i, T_j)$}
    \State $\mathbf{T} = \text{exp}\left(-\frac{\mathbf{C}}{\tau}\right)$ 
    \State $\mathbf{T} = \frac{s}{\left(\mathbf{1}_{T_i}\right)^{\top} \cdot \mathbf{T} \cdot \mathbf{1}_{T_j}} \mathbf{T}$
    
    \For{$i=1$ to $N_{\text{iter}}$} 
        \State $\mathbf{p}_{a} = \min \left(\frac{\mathbf{a}}{\mathbf{T}\mathbf{1}_{T_j}}, \mathbf{1}_{T_i}\right)$, $\mathbf{T}_{a} = \text{diag}\left(\mathbf{p}_{a}\right) \cdot \mathbf{T}$
        \State $\mathbf{p}_{b} = \min \left(\frac{\mathbf{b}}{\mathbf{T}_{a}^{\top}\mathbf{1}_{T_i}}, \mathbf{1}_{T_j}\right)$, $\mathbf{T}_{b} = \text{diag}\left(\mathbf{p}_{b}\right) \cdot \mathbf{T}_{a }$
        \State $\mathbf{T} = \frac{s}{\left(\mathbf{1}_{T_i}\right)^{\top} \cdot \mathbf{T} \cdot \mathbf{1}_{N_L}} \mathbf{T}_{b}$
    \EndFor 
    \State $d_{\text{OT}} = \min\left(d_{\text{OT}}, \;\sum\limits_{k=1}^{T_i} \sum\limits_{l=1}^{T_j} \mathbf{T}_{k,l} \mathbf{C}_{k,l}\right)$ 
    \EndFor \\
\Return $d_{\text{OT}}$
\end{algorithmic}}
\end{algorithm}
\section{Experiments}
{\renewcommand{\arraystretch}{1.1}
\begin{table*}[t]
\centering
\resizebox{0.7\linewidth}{!}{
\begin{tabular}{l|ccc|ccc}
\hline
\multicolumn{1}{c|}{\multirow{2}{*}{\textbf{Method}}} & \multicolumn{3}{c|}{\textbf{vIoU threshold = 0.5}}       & \multicolumn{3}{c}{\textbf{vIoU threshold = 0.1}}       \\
\multicolumn{1}{c|}{}                                 & \textbf{R/mR@20} & \textbf{R/mR@50} & \textbf{R/mR@100} & \textbf{R/mR@20} & \textbf{R/mR@50} & \textbf{R/mR@100} \\ \hline
IPS+T - Vanilla                                      & 3.04 / 1.35      & 4.61 / 2.94      & 5.56 / 3.33       & 8.28 / 5.68      & 14.47 / 9.92     & 18.24 / 11.84     \\
IPS+T - Handcrafted filter                           & 2.52 / 1.72      & 3.77 / 2.36      & 4.72 / 2.79       & 8.07 / 5.61      & 13.42 / 8.27     & 16.46 / 10.11     \\
IPS+T - Transformer                                  & 3.88 / 2.81      & 5.66 / 4.12      & 6.18 / 4.44       & 9.01 / 6.69      & 14.88 / 11.28    & 17.51 / 13.20     \\
IPS+T - Convolution                                  & 3.88 / 2.55      & 5.24 / 3.29      & 6.71 / 5.36       & 10.06 / 8.98     & 14.99 / 12.21    & 18.13 / 15.47     \\
Ours - Transformer                                               &    \underline{3.98} / \underline{2.98}					              &      \underline{5.97} / \underline{4.20}            &      \underline{7.44} / \underline{5.15}             &          \underline{10.59} / \underline{9.56}        &     \underline{16.98} / \underline{12.39}             &      \underline{22.33} / \underline{17.47}             \\ 
Ours - Convolution                                                &   \textbf{4.51} / \textbf{3.56}				              &             \textbf{6.08} / \textbf{4.38}     &      \textbf{7.76} / \textbf{5.86}	             &     \textbf{11.43} / \textbf{9.57}             &    \textbf{17.30} / \textbf{13.13}              &       \textbf{22.85} / \textbf{17.48}            \\ \hline
VPS - Vanilla                                        & 0.21 / 0.10      & 0.21 / 0.10      & 0.31 / 0.18       & 6.29 / 3.04      & 9.64 / 6.74      & 12.89 / 9.60      \\
VPS - Handcrafted filter                             & 0.42 / 0.13      & 0.52 / 0.50      & 0.94 / 0.92       & 5.24 / 2.84      & 7.65 / 7.14      & 9.64 / 8.22       \\
VPS - Transformer                                    & 0.42 / 0.61      & 0.73 / 0.76      & 1.05 / 0.92       & 6.50 / 5.75      & 9.64 / 8.25      & 12.26 / 9.51      \\
VPS - Convolution                                    & 0.42 / 0.25      & 0.63 / 0.67      & 0.63 / 0.67       & 8.07 / 7.84      & 11.01 / 9.78     & 12.89 / 10.77     \\
Ours - Transformer                                                &    \underline{0.63} / \underline{0.83}              &    \underline{1.05} / \underline{0.76}              &      \underline{1.05} / \underline{0.76}             &     \underline{6.71} / \underline{6.94}             &   \underline{10.27} / \underline{8.68}               &     \underline{13.42} / \underline{12.09}              \\ 
Ours - Convolution                                               &   \textbf{0.84} / \textbf{0.98}					               &         \textbf{1.26} / \textbf{1.22}         &        \textbf{1.26} / \textbf{1.22}           &       \textbf{8.18} / \textbf{8.00}           &     \textbf{12.90} / \textbf{11.47}             &      \textbf{14.22} / \textbf{13.59}             \\ \hline
\end{tabular}}
\vspace{-5pt}
\caption{Experimental results on the OpenPVSG dataset.}
\label{tab:exp_openpvsg}
\end{table*}}

{\renewcommand{\arraystretch}{1.1}
\begin{table*}[t]
\centering
\resizebox{0.7\linewidth}{!}{
\begin{tabular}{l|l|ccc|ccc}
\hline
\multirow{2}{*}{\textbf{Input type}} & \multicolumn{1}{c|}{\multirow{2}{*}{\textbf{Method}}} & \multicolumn{3}{c|}{\textbf{PSG4D-GTA}}                  & \multicolumn{3}{c}{\textbf{PSG4D-HOI}}                  \\ 
                                     &                                  & \textbf{R/mR@20} & \textbf{R/mR@50} & \textbf{R/mR@100} & \textbf{R/mR@20} & \textbf{R/mR@50} & \textbf{R/mR@100} \\ \hline
\multirow{3}{*}{Point cloud videos}  & 3DSGG                            & 1.48 / 0.73      & 2.16 / 0.79      & 2.92 / 0.85       & 3.46 / 2.19      & 3.15 / 2.47      & 4.96 / 2.84       \\
                                     & PSG4DFormer                      & 4.33 / 2.10      & 4.83 / 2.93      & 5.22 / 3.13       & 5.36 / 3.10      & 5.61 / 3.95      & 6.76 / 4.17       \\
                                     & Ours                             & \textbf{5.88} / \textbf{3.45}      & \textbf{6.31} / \textbf{3.70}      & \textbf{7.31} / \textbf{4.70}       & \textbf{7.28} / \textbf{5.09}      & \textbf{7.62} / \textbf{6.49}      & \textbf{9.18} / \textbf{6.85}       \\ \hline
\multirow{3}{*}{RGB-D videos}        & 3DSGG                            & 2.29 / 0.92      & 2.46 / 1.01      & 3.81 / 1.45       & 4.23 / 2.19      & 4.47 / 2.31      & 4.86 / 2.41       \\
                                     & PSG4DFormer                      & 6.68 / 3.31      & 7.17 / 3.85      & 7.22 / 4.02       & 5.62 / 3.65      & 6.16 / 4.16      & 6.28 / 4.97       \\
                                     & Ours                             & \textbf{9.07} / \textbf{5.52}      & \textbf{9.73} / \textbf{6.32}      & \textbf{9.73} / \textbf{6.32}       & \textbf{7.63} / \textbf{6.09}      & \textbf{8.36} / \textbf{6.94}      & \textbf{8.53} / \textbf{8.29} \\ \hline      
\end{tabular}}
\caption{Experimental results on both PSG4D-GTA and PSG4D-HOI groups of PSG4D dataset.}
\label{tab:exp_psg4d}
\vspace{-10pt}
\end{table*}}

{\renewcommand{\arraystretch}{1.1}
\begin{table}[t]
\centering
\resizebox{0.85\linewidth}{!}{
\begin{tabular}{l|ccc}
\hline
\multicolumn{1}{c|}{\textbf{Method}} & \textbf{R/mR@20} & \textbf{R/mR@50} & \textbf{R/mR@100} \\ \hline
w/o shuffle-based                   & 4.41 / 3.43      & 5.90 / 4.24      & 7.30 / 5.79      \\
w/o triplet-based                   & 4.44 / 3.50      & 6.02 / 4.28      & 7.36 / 5.83       \\
Ours                                &   \textbf{4.51} / \textbf{3.56}               &   \textbf{6.08} / \textbf{4.38}               &    \textbf{7.44} / \textbf{5.86}               \\ \hline
\end{tabular}}
\caption{Ablation results for contrastive learning approaches on OpenPVSG dataset. We adopt the vIoU threshold of 0.5.}
\label{tab:exp_ablation_contrastive_openpvsg}
\vspace{-10pt}
\end{table}}

{\renewcommand{\arraystretch}{1.1}
\begin{table}[t]
\centering
\resizebox{0.95\linewidth}{!}{
\begin{tabular}{l|ccc}
\hline
\multicolumn{1}{c|}{\textbf{Tube relation quantification}} & \textbf{R/mR@20} & \textbf{R/mR@50} & \textbf{R/mR@100} \\ \hline
Pooling - Cosine similarity                   & 4.44 / 3.40      & 6.04 / 4.36      & 7.36 / 5.84       \\
Pooling - L2                   & 4.36 / 3.37      & 3.77 / 5.95      & 7.29 / 5.80       \\
Optimal transport                                &    \textbf{4.51} / \textbf{3.56}              &    \textbf{6.08} / \textbf{4.38}              &  \textbf{7.44} / \textbf{5.86}                 \\ \hline
\end{tabular}}
\caption{Ablation results for mask tube relation quantification method between mask tubes on OpenPVSG dataset.}
\label{tab:exp_ablation_ot_distance_openpvsg}
\end{table}}

{\renewcommand{\arraystretch}{1.1}
\begin{table*}[t]
\centering
\resizebox{0.8\linewidth}{!}{
\begin{tabular}{c|l|ccc|ccc}
\hline
\multirow{2}{*}{\textbf{Input type}} & \multicolumn{1}{c|}{\multirow{2}{*}{\textbf{Method}}} & \multicolumn{3}{c|}{\textbf{PSG4D-GTA}}                  & \multicolumn{3}{c}{\textbf{PSG4D-HOI}}                  \\
                                     &                                 & \textbf{R/mR@20} & \textbf{R/mR@50} & \textbf{R/mR@100} & \textbf{R/mR@20} & \textbf{R/mR@50} & \textbf{R/mR@100} \\ \hline
\multirow{3}{*}{Point cloud videos}  & w/o shuffle-based                                    & 5.56 / 2.92      & 5.57 / 2.98      & 6.51 / 4.36      & 6.56 / 4.29      & 6.98 / 6.25      & 8.76 / 6.43       \\
& w/o triplet-based                                    & 5.77 / 2.93      & 5.59 / 3.26      & 6.53 / 4.39       & 6.67 / 4.85     & 7.52 / 6.31      & 8.84 / 6.43      \\
                                     & Ours                                                 & \textbf{5.88} / \textbf{3.45}      & \textbf{6.31} / \textbf{3.70}      & \textbf{7.31} / \textbf{4.70}       & \textbf{7.28} / \textbf{5.09}      & \textbf{7.62} / \textbf{6.49}      & \textbf{9.18} / \textbf{6.85}        \\ \hline
\multirow{3.2}{*}{RGB-D videos}     & w/o shuffle-based                                    & 8.35 / 5.34      & 8.76 / 5.68      & 8.88 / 5.53       & 7.00 / 5.53      & 7.51 / 6.02      & 7.56 / 7.42       \\
                              & w/o triplet-based                                    & 9.00 / 5.46      & 9.71 / 5.95      & 9.63 / 5.82       & 7.12 / 6.03      & 8.31 / 6.51      & 8.24 / 7.95       \\
                                     & Ours                                                 & \textbf{9.07} / \textbf{5.52}      & \textbf{9.73} / \textbf{6.32}      & \textbf{9.73} / \textbf{6.32}       & \textbf{7.63} / \textbf{6.09}      & \textbf{8.36} / \textbf{6.94}      & \textbf{8.53} / \textbf{8.29}      \\ \hline
\end{tabular}}
\caption{Ablation results for contrastive learning approaches on PSG4D dataset.}
\label{tab:exp_ablation_contrastive_psg4d}
\end{table*}}

{\renewcommand{\arraystretch}{1.1}
\begin{table*}[t]
\centering
\resizebox{0.8\linewidth}{!}{
\begin{tabular}{c|l|ccc|ccc}
\hline
\multirow{2}{*}{\textbf{Input type}} & \multicolumn{1}{c|}{\multirow{2}{*}{\textbf{Tube relation quantification}}} & \multicolumn{3}{c|}{\textbf{PSG4D-GTA}}                  & \multicolumn{3}{c}{\textbf{PSG4D-HOI}}                  \\
                                     &                                 & \textbf{R/mR@20} & \textbf{R/mR@50} & \textbf{R/mR@100} & \textbf{R/mR@20} & \textbf{R/mR@50} & \textbf{R/mR@100} \\ \hline
\multirow{3}{*}{Point cloud videos}  & Pooling - Cosine similarity                                   & 5.76 / 2.87     & 6.02 / 3.62      & 6.84 / 4.11       & 7.24 / 4.45      & 7.44 / 6.27      & 8.20 / 6.64       \\
                                     & Pooling - L2                                  & 5.46 / 2.78      & 5.38 / 3.39      & 6.51 / 3.86       & 6.72 / 4.11      & 6.74 / 6.05      & 7.96 / 6.11       \\
                                     & Optimal transport                                                 & \textbf{5.88} / \textbf{3.45}      & \textbf{6.31} / \textbf{3.70}      & \textbf{7.31} / \textbf{4.70}       & \textbf{7.28} / \textbf{5.09}      & \textbf{7.62} / \textbf{6.49}      & \textbf{9.18} / \textbf{6.85}       \\ \hline
\multirow{3.2}{*}{RGB-D videos}        & Pooling - Cosine similarity                                    & 9.03 / 5.37     & 9.47 / 5.86      & 9.70 / 6.02       & 7.36 / 5.43      & 7.93 / 6.70     & 8.06 / 7.42      \\
                                     & Pooling - L2                                  & 8.89 / 4.70      & 8.90 / 5.41      & 9.08 / 5.78       & 6.65 / 5.26      & 7.74 / 6.29      & 7.95 / 7.39       \\
                                     & Optimal transport                                                 & \textbf{9.07} / \textbf{5.52}      & \textbf{9.73} / \textbf{6.32}      & \textbf{9.73} / \textbf{6.32}       & \textbf{7.63} / \textbf{6.09}      & \textbf{8.36} / \textbf{6.94}      & \textbf{8.53} / \textbf{8.29}     \\ \hline
\end{tabular}}
\caption{Ablation results for mask tube relation quantification method between mask tubes on PSG4D dataset.}
\label{tab:exp_ablation_ot_distance_psg4d}
\vspace{-10pt}
\end{table*}}

We conduct comprehensive experiments to evaluate the effectiveness of our motion-aware contrastive framework. We first describe the experiment settings, covering the evaluation datasets, evaluation metrics, baseline methods, and implementation details. Next, we present quantitative results of our method, then provide ablation study and careful analysis to explore properties of our motion-aware contrastive framework. Eventually, we conduct qualitative analysis to concretely examine its behavior.

\vspace{-5pt}
\subsection{Experiment Settings}
\noindent\textbf{Datasets.} We assess the effectiveness of our method on natural and 4D video inputs. The corresponding dataset to each input type is as follows:
\begin{itemize}
    \item \textbf{Open-domain Panoptic video scene graph generation (OpenPVSG)} \citep{yang2023panoptic}: OpenPVSG consists of scene graphs and associated segmentation masks with respect to subject and object nodes in the scene graph. The dataset comprises 400 videos, including 289 third-person videos from ViDOR \citep{shang2019annotating}, 111 egocentric videos from Epic-Kitchens \citep{damen2022epic} and Ego4D \citep{grauman2022ego4d}. 

    \item \textbf{Panoptic scene graph generation for 4D (PSG4D)} \citep{yang20244d}: The PSG4D dataset is divided into two groups, \textit{i.e.} PSG4D-GTA and PSG4D-HOI. PSG4D-GTA comprises 67 third-view videos with an average length of 84 seconds, 35 object categories, and 43 relationship categories. On the contrary, PSG4D-HOI contains 2,973 videos from an egocentric perspective, whose average duration is 20 seconds. The PSG4D-HOI's videos are mostly related to indoor scenes, covering 46 object categories and 15 relationship categories. 
\end{itemize}

\noindent\textbf{Evaluation metrics.} We use the recall at $K$ (R@$K$) and mean recall at $K$ (mR@$K$) metrics, which are standard metrics used in scene graph generation tasks. Both R@$K$ and mR@$K$ consider the top-$K$ triplets predicted by the panoptic scene graph generation model. A successful recall of a predicted triplet must satisfy the following criteria: 1) correct category labels for the subject, object, and predicate; 2) a volume Intersection over Union (vIoU) greater than or equal to 0.5 between the predicted mask tubes and the groundtruth tubes. For extensive comparison, we also report results with the vIoU threshold of 0.1. 

\noindent\textbf{Baseline methods.} We compare our method with a comprehensive list of baseline approaches for temporal panoptic scene graph generation: (i) \textbf{IPS+T - Vanilla} \citep{yang2023panoptic} uses image panoptic segmentation (IPS) model with a tracker for segmentation, and fully-connected layers to separately encode temporal states of entity mask tubes; (ii) \textbf{IPS+T - Handcrafted filter} \citep{yang2023panoptic} uses image panoptic segmentation (IPS) model with a tracker for segmentation, and a manually-designed kernel to encode entity mask tubes; (iii) \textbf{IPS+T - Convolution} \citep{yang2023panoptic} uses image panoptic segmentation (IPS) model with a tracker for segmentation, and learnable convolutional layers to encode entity mask tubes; (iv) \textbf{IPS+T - Transformer} \citep{yang2023panoptic} uses image panoptic segmentation model (IPS) with a tracker for segmentation, and Transformer-based encoder with self-attention layers to encode entity mask tubes; (v) \textbf{VPS - Vanilla} \citep{yang2023panoptic} is similar to IPS+T - Vanilla, but uses video panoptic segmentation (VPS) model for panoptic segmentation; (vi) \textbf{VPS - Handcrafted filter} \citep{yang2023panoptic} is similar to IPS+T - Handcrafter filter, but uses video panoptic segmentation (VPS) model for  segmentation; (vii) \textbf{VPS - Convolution} \citep{yang2023panoptic} is similar to IPS+T - Convolution, but uses video panoptic segmentation (VPS) model for segmentation; (viii) \textbf{VPS - Transformer} \citep{yang2023panoptic} is similar to IPS+T - Transformer, but uses video panoptic segmentation (VPS) model for segmentation; (ix) \textbf{3D-SGG} \citep{wald2020learning} is based on PointNet \citep{qi2017pointnet} and graph convolutional network \citep{kipf2016semi} but neglects the depth dimension and generates panoptic scene graphs for 4D video inputs; (x) \textbf{PSG4DFormer} \citep{yang20244d} is a specialized model for 4D inputs, using Mask2Former \citep{cheng2022masked} for segmentation and a spatial-temporal Transformer to encode object mask tubes for relation classification.

\noindent\textbf{Implementation details.} For fair comparison, we experiment our contrastive framework with both IPS+T and VPS as segmentation module for panoptic video scene graph generation. In the former case, we leverage the UniTrack tracker \citep{wang2021different} combined with Mask2Former model \citep{cheng2022masked}, which is initialized from the best-performing COCO-pretrained weights and fine-tuned for 8 epochs using AdamW optimizer with a batch size of 32, learning rate of 0.0001, weight decay of 0.05, and gradient clipping with a max L2 norm of 0.01. In the latter case, we utilize Video K-Net \citep{li2022video}, also initialized from COCO-pretrained weights and fine-tuned with the same strategy as IPS+T. In the relation classification step, we conduct fine-tuning with a batch size of 32, employing the Adam optimizer with a learning rate of 0.001. For 4D panoptic scene graph generation, we adopt the PSG4DFormer baseline. To work with RGB-D and point cloud videos, we use an ImageNet pretrained on ResNet-101 \citep{russakovsky2015imagenet} and the DKNet \citep{wu20223d} as the visual encoder, respectively. We fine-tune the segmentation module for RGB-D and point cloud videos for 12 and 200 epochs, respectively. We use additional 100 epochs to train the relation classification module. Based on validation, we adopt a threshold $\gamma = 9.0$ and a margin $\alpha = 10.0$. We set the maximum number of iterations $N_{\text{iter}}$ to 1,000. 

\begin{figure}[t]
\centering
\includegraphics[width=0.6\linewidth]{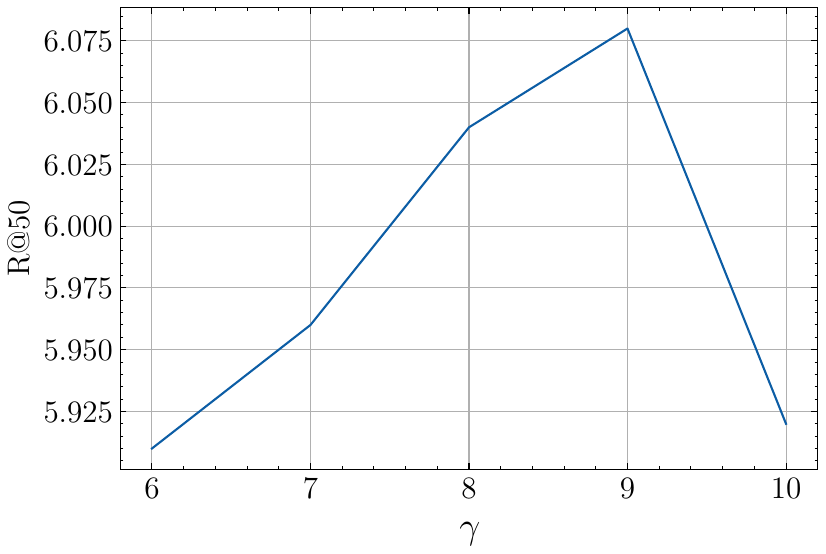}
\vspace{-5pt}
\caption{Ablation results on threshold $\gamma$.}
\label{fig:ablation_gamma}
\end{figure}

\vspace{-5pt}
\subsection{Main Results}
\noindent\textbf{Results on OpenPVSG.} As shown in Table \ref{tab:exp_openpvsg}, we substantially outperform both IPS+T - Convolution and IPS+T - Transformer when we use IPS+T for segmentation. In particular, using a higher vIoU threshold to filter out inaccurate segmentation, we surpass IPS+T - Transformer by 1.3/0.7 points of R/mR@100, while surpassing IPS+T - Convolution by 0.8/1.1 points of R/mR@50. In addition, for a less strict vIoU threshold, we outperform IPS+T - Transformer by 1.6/2.9 points of R/mR@20, and IPS+T - Convolution by 2.3/0.9 points of R/mR@50. These results demonstrate that our method makes a propitious contribution to temporal panoptic scene graph generation, not only to popular but also to unpopular relation classes. 

\noindent\textbf{Results on PSG4D.} Table \ref{tab:exp_psg4d} shows that our method also achieves significantly higher performance than the PSG4DFormer model. Specifically, when working with point cloud videos, on PSG4D-GTA, we outperform the baseline method by 1.6/1.4 points. Analogously, on PSG4D-HOI, we outperform PSG4DFormer by 2.0/2.5 points of R/mR@50. These results indicate that our framework bears a valuable impact to both egocentric and third-view videos. We hypothesize that both video types consist of dynamic actions among objects whose mask tube representations should be polished. In addition, when working with RGB-D videos, on PSG4D-GTA, we enhance the baseline method by 2.4/2.2 points of R/mR@20,. Furthermore, on PSG4D-HOI, our motion-aware contrastive learning also considerably refines PSG4DFormer by 2.0/2.4 points of R/mR@20. 

Such results have verified the generalizability of our motion-aware contrastive framework over natural, point cloud, and RGB-D videos.

\vspace{-5pt}
\subsection{Ablation Study}
\noindent\textbf{Effect of the contrastive components.} We evaluate our framework without the assistance of either the shuffle-based or the triplet-based contrastive objective. As shown in Table \ref{tab:exp_ablation_contrastive_openpvsg} and \ref{tab:exp_ablation_contrastive_psg4d}, the performance degrades when we both remove shuffle-based and triplet-based contrastive approaches. In addition, triplet-based contrastive learning plays a more fundamental role than the shuffle-based one. We hypothesize that shuffle-based contrastive learning is better at focusing on motion semantics than triplet-based one.

\noindent\textbf{Effect of selecting strong-motion tubes.} We evaluate the impact of our strategy to filter out weak motion tubes. In Figure \ref{fig:ablation_gamma}, we observe a performance boost when we increase the threshold to select mask tubes with strong motion. However, further elevating the threshold results in performance degradation, since there are more mask tubes eliminated, thus limiting the effect of our motion-aware contrastive framework. 

\noindent\textbf{Effect of optimal transport distance.} In this ablation, we compare various strategies to calculate the similarity between two mask tubes. Results in Table \ref{tab:exp_ablation_ot_distance_openpvsg} and \ref{tab:exp_ablation_ot_distance_psg4d} show that the proposed optimal transport achieves much higher performance for both natural and 4D video inputs. We conjecture that other method such as pooling then cosine similarity or L2 neglects the temporal or flattens the motion nature of the entity mask tubes, thus reducing the effectiveness.

\vspace{-5pt}
\subsection{Qualitative Analysis}
We visualize examples processed by the state-of-the-art models and ours in Figure \ref{fig:qualitative_example}. As can be observed, our model successfully produces mask tubes overlapping with the groundtruth, and importantly predicts the correct relations of the subject-object pairs. On the other hand, baseline models tend to prefer more static relations, since during training they do not explicitly focus on motion-sensitive features. Statistics in Figure \ref{fig:r@50} also substantiate our proposition, in which we achieve considerably higher recalls for dynamic relations than baseline approaches.
\vspace{-5pt}
\section{Conclusion}
In this paper, we propose a motion-aware contrastive learning framework for temporal panoptic scene graph generation. In our framework, we learn close representations for temporal masks of similar entities that exhibit common relations. Moreover, we separate temporal masks from their shuffled version, and also separate temporal masks of different subject-relation-object triplets. To quantify the relationship among temporal masks in the proposed contrastive framework, we utilize optimal transport to preserve the temporal nature among temporal entity masks. Extensive experiments substantiate the effectiveness of our framework for both natural and 4D videos.
\section*{Acknowledgements}
This research/project is supported by the National Research Foundation, Singapore under its AI Singapore Programme (AISG Award No: AISG3-PhD-2023-08-051T). Thong Nguyen is supported by a Google Ph.D. Fellowship in Natural Language Processing.

\bibliography{aaai25}

@article{ma2022sqa3d,
  title={Sqa3d: Situated question answering in 3d scenes},
  author={Ma, Xiaojian and Yong, Silong and Zheng, Zilong and Li, Qing and Liang, Yitao and Zhu, Song-Chun and Huang, Siyuan},
  journal={arXiv preprint arXiv:2210.07474},
  year={2022}
}

@article{driess2023palm,
  title={Palm-e: An embodied multimodal language model},
  author={Driess, Danny and Xia, Fei and Sajjadi, Mehdi SM and Lynch, Corey and Chowdhery, Aakanksha and Ichter, Brian and Wahid, Ayzaan and Tompson, Jonathan and Vuong, Quan and Yu, Tianhe and others},
  journal={arXiv preprint arXiv:2303.03378},
  year={2023}
}

@article{raychaudhuri2023reduce,
  title={Reduce, reuse, recycle: Modular multi-object navigation},
  author={Raychaudhuri, Sonia and Campari, Tommaso and Jain, Unnat and Savva, Manolis and Chang, Angel X},
  journal={arXiv preprint arXiv:2304.03696},
  volume={2},
  year={2023}
}

@inproceedings{cheng2022masked,
  title={Masked-attention mask transformer for universal image segmentation},
  author={Cheng, Bowen and Misra, Ishan and Schwing, Alexander G and Kirillov, Alexander and Girdhar, Rohit},
  booktitle={Proceedings of the IEEE/CVF conference on computer vision and pattern recognition},
  pages={1290--1299},
  year={2022}
}

@inproceedings{li2023tube,
  title={Tube-Link: A flexible cross tube framework for universal video segmentation},
  author={Li, Xiangtai and Yuan, Haobo and Zhang, Wenwei and Cheng, Guangliang and Pang, Jiangmiao and Loy, Chen Change},
  booktitle={Proceedings of the IEEE/CVF International Conference on Computer Vision},
  pages={13923--13933},
  year={2023}
}

@article{li2023transformer,
  title={Transformer-based visual segmentation: A survey},
  author={Li, Xiangtai and Ding, Henghui and Yuan, Haobo and Zhang, Wenwei and Pang, Jiangmiao and Cheng, Guangliang and Chen, Kai and Liu, Ziwei and Loy, Chen Change},
  journal={arXiv preprint arXiv:2304.09854},
  year={2023}
}

@inproceedings{sudhakaran2023vision,
  title={Vision relation transformer for unbiased scene graph generation},
  author={Sudhakaran, Gopika and Dhami, Devendra Singh and Kersting, Kristian and Roth, Stefan},
  booktitle={Proceedings of the IEEE/CVF International Conference on Computer Vision},
  pages={21882--21893},
  year={2023}
}

@inproceedings{nag2023unbiased,
  title={Unbiased scene graph generation in videos},
  author={Nag, Sayak and Min, Kyle and Tripathi, Subarna and Roy-Chowdhury, Amit K},
  booktitle={Proceedings of the IEEE/CVF Conference on Computer Vision and Pattern Recognition},
  pages={22803--22813},
  year={2023}
}

@inproceedings{wang2024oed,
  title={OED: Towards One-stage End-to-End Dynamic Scene Graph Generation},
  author={Wang, Guan and Li, Zhimin and Chen, Qingchao and Liu, Yang},
  booktitle={Proceedings of the IEEE/CVF Conference on Computer Vision and Pattern Recognition},
  pages={27938--27947},
  year={2024}
}

@inproceedings{li2022dynamic,
  title={Dynamic scene graph generation via anticipatory pre-training},
  author={Li, Yiming and Yang, Xiaoshan and Xu, Changsheng},
  booktitle={Proceedings of the IEEE/CVF conference on computer vision and pattern recognition},
  pages={13874--13883},
  year={2022}
}

@inproceedings{yang2023panoptic,
  title={Panoptic video scene graph generation},
  author={Yang, Jingkang and Peng, Wenxuan and Li, Xiangtai and Guo, Zujin and Chen, Liangyu and Li, Bo and Ma, Zheng and Zhou, Kaiyang and Zhang, Wayne and Loy, Chen Change and others},
  booktitle={Proceedings of the IEEE/CVF Conference on Computer Vision and Pattern Recognition},
  pages={18675--18685},
  year={2023}
}

@article{yang20244d,
  title={4d panoptic scene graph generation},
  author={Yang, Jingkang and Cen, Jun and Peng, Wenxuan and Liu, Shuai and Hong, Fangzhou and Li, Xiangtai and Zhou, Kaiyang and Chen, Qifeng and Liu, Ziwei},
  journal={Advances in Neural Information Processing Systems},
  volume={36},
  year={2024}
}

@inproceedings{yang2022panoptic,
  title={Panoptic scene graph generation},
  author={Yang, Jingkang and Ang, Yi Zhe and Guo, Zujin and Zhou, Kaiyang and Zhang, Wayne and Liu, Ziwei},
  booktitle={European Conference on Computer Vision},
  pages={178--196},
  year={2022},
  organization={Springer}
}

@inproceedings{zhao2023textpsg,
  title={Textpsg: Panoptic scene graph generation from textual descriptions},
  author={Zhao, Chengyang and Shen, Yikang and Chen, Zhenfang and Ding, Mingyu and Gan, Chuang},
  booktitle={Proceedings of the IEEE/CVF International Conference on Computer Vision},
  pages={2839--2850},
  year={2023}
}

@article{wang2023taking,
  title={Taking a closer look at visual relation: Unbiased video scene graph generation with decoupled label learning},
  author={Wang, Wenqing and Luo, Yawei and Chen, Zhiqing and Jiang, Tao and Yang, Yi and Xiao, Jun},
  journal={IEEE Transactions on Multimedia},
  year={2023},
  publisher={IEEE}
}

@article{zhou2022context,
  title={Context-aware mixture-of-experts for unbiased scene graph generation},
  author={Zhou, Liguang and Zhou, Yuhongze and Lam, Tin Lun and Xu, Yangsheng},
  journal={arXiv preprint arXiv:2208.07109},
  year={2022}
}

@article{chen2020graph,
  title={Graph convolutional network with structure pooling and joint-wise channel attention for action recognition},
  author={Chen, Yuxin and Ma, Gaoqun and Yuan, Chunfeng and Li, Bing and Zhang, Hui and Wang, Fangshi and Hu, Weiming},
  journal={Pattern Recognition},
  volume={103},
  pages={107321},
  year={2020},
  publisher={Elsevier}
}

@article{liu2022spatial,
  title={Spatial focus attention for fine-grained skeleton-based action tasks},
  author={Liu, Kaiyuan and Li, Yunheng and Xu, Yuanfeng and Liu, Shuai and Liu, Shenglan},
  journal={IEEE Signal Processing Letters},
  volume={29},
  pages={1883--1887},
  year={2022},
  publisher={IEEE}
}

@inproceedings{zhou2023learning,
  title={Learning discriminative representations for skeleton based action recognition},
  author={Zhou, Huanyu and Liu, Qingjie and Wang, Yunhong},
  booktitle={Proceedings of the IEEE/CVF Conference on Computer Vision and Pattern Recognition},
  pages={10608--10617},
  year={2023}
}

@article{wang2021different,
  title={Do different tracking tasks require different appearance models?},
  author={Wang, Zhongdao and Zhao, Hengshuang and Li, Ya-Li and Wang, Shengjin and Torr, Philip and Bertinetto, Luca},
  journal={Advances in Neural Information Processing Systems},
  volume={34},
  pages={726--738},
  year={2021}
}

@article{xiao2021modist,
  title={Modist: Motion distillation for self-supervised video representation learning},
  author={Xiao, Fanyi and Tighe, Joseph and Modolo, Davide},
  journal={arXiv preprint arXiv:2106.09703},
  volume={3},
  year={2021}
}

@article{sobel2022sobel,
  title={Sobel-feldman operator},
  author={Sobel, Irwin and Duda, R and Hart, P and Wiley, John},
  journal={Preprint at https://www. researchgate. net/profile/Irwin-Sobel/publication/285159837. Accessed},
  volume={20},
  year={2022}
}

@inproceedings{shang2019annotating,
  title={Annotating objects and relations in user-generated videos},
  author={Shang, Xindi and Di, Donglin and Xiao, Junbin and Cao, Yu and Yang, Xun and Chua, Tat-Seng},
  booktitle={Proceedings of the 2019 on International Conference on Multimedia Retrieval},
  pages={279--287},
  year={2019}
}

@article{damen2022epic,
  title={Epic-kitchens-100},
  author={Damen, Dima and Doughty, Hazel and Farinella, Giovanni Maria and Furnari, Antonino and Kazakos, Evangelos and Ma, Jian and Moltisanti, Davide and Munro, Jonathan and Perrett, Toby and Price, Will and others},
  journal={International Journal of Computer Vision},
  volume={130},
  pages={33--55},
  year={2022}
}

@inproceedings{grauman2022ego4d,
  title={Ego4d: Around the world in 3,000 hours of egocentric video},
  author={Grauman, Kristen and Westbury, Andrew and Byrne, Eugene and Chavis, Zachary and Furnari, Antonino and Girdhar, Rohit and Hamburger, Jackson and Jiang, Hao and Liu, Miao and Liu, Xingyu and others},
  booktitle={Proceedings of the IEEE/CVF Conference on Computer Vision and Pattern Recognition},
  pages={18995--19012},
  year={2022}
}

@inproceedings{wald2020learning,
  title={Learning 3d semantic scene graphs from 3d indoor reconstructions},
  author={Wald, Johanna and Dhamo, Helisa and Navab, Nassir and Tombari, Federico},
  booktitle={Proceedings of the IEEE/CVF Conference on Computer Vision and Pattern Recognition},
  pages={3961--3970},
  year={2020}
}

@inproceedings{qi2017pointnet,
  title={Pointnet: Deep learning on point sets for 3d classification and segmentation},
  author={Qi, Charles R and Su, Hao and Mo, Kaichun and Guibas, Leonidas J},
  booktitle={Proceedings of the IEEE conference on computer vision and pattern recognition},
  pages={652--660},
  year={2017}
}

@article{kipf2016semi,
  title={Semi-supervised classification with graph convolutional networks},
  author={Kipf, Thomas N and Welling, Max},
  journal={arXiv preprint arXiv:1609.02907},
  year={2016}
}

@inproceedings{wu20223d,
  title={3d instances as 1d kernels},
  author={Wu, Yizheng and Shi, Min and Du, Shuaiyuan and Lu, Hao and Cao, Zhiguo and Zhong, Weicai},
  booktitle={European Conference on Computer Vision},
  pages={235--252},
  year={2022},
  organization={Springer}
}

@inproceedings{li2022video,
  title={Video k-net: A simple, strong, and unified baseline for video segmentation},
  author={Li, Xiangtai and Zhang, Wenwei and Pang, Jiangmiao and Chen, Kai and Cheng, Guangliang and Tong, Yunhai and Loy, Chen Change},
  booktitle={Proceedings of the IEEE/CVF Conference on Computer Vision and Pattern Recognition},
  pages={18847--18857},
  year={2022}
}

@article{russakovsky2015imagenet,
  title={Imagenet large scale visual recognition challenge},
  author={Russakovsky, Olga and Deng, Jia and Su, Hao and Krause, Jonathan and Satheesh, Sanjeev and Ma, Sean and Huang, Zhiheng and Karpathy, Andrej and Khosla, Aditya and Bernstein, Michael and others},
  journal={International journal of computer vision},
  volume={115},
  pages={211--252},
  year={2015},
  publisher={Springer}
}

@inproceedings{dong2024memflow,
  title={MemFlow: Optical Flow Estimation and Prediction with Memory},
  author={Dong, Qiaole and Fu, Yanwei},
  booktitle={Proceedings of the IEEE/CVF Conference on Computer Vision and Pattern Recognition},
  pages={19068--19078},
  year={2024}
}

@inproceedings{davtyan2023efficient,
  title={Efficient video prediction via sparsely conditioned flow matching},
  author={Davtyan, Aram and Sameni, Sepehr and Favaro, Paolo},
  booktitle={Proceedings of the IEEE/CVF International Conference on Computer Vision},
  pages={23263--23274},
  year={2023}
}

@article{shen2024patch,
  title={Patch Spatio-Temporal Relation Prediction for Video Anomaly Detection},
  author={Shen, Hao and Shi, Lu and Xu, Wanru and Cen, Yigang and Zhang, Linna and An, Gaoyun},
  journal={arXiv preprint arXiv:2403.19111},
  year={2024}
}

@article{ren2024arvideo,
  title={ARVideo: Autoregressive Pretraining for Self-Supervised Video Representation Learning},
  author={Ren, Sucheng and Zhu, Hongru and Wei, Chen and Li, Yijiang and Yuille, Alan and Xie, Cihang},
  journal={arXiv preprint arXiv:2405.15160},
  year={2024}
}

@article{nguyen2024meta,
  title={Meta-optimized Angular Margin Contrastive Framework for Video-Language Representation Learning},
  author={Nguyen, Thong and Bin, Yi and Wu, Xiaobao and Dong, Xinshuai and Hu, Zhiyuan and Le, Khoi and Nguyen, Cong-Duy and Ng, See-Kiong and Tuan, Luu Anh},
  journal={arXiv preprint arXiv:2407.03788},
  year={2024}
}

@article{liu2024contrastive,
  title={Contrastive Language Video Time Pre-training},
  author={Liu, Hengyue and Min, Kyle and Valdez, Hector A and Tripathi, Subarna},
  journal={arXiv preprint arXiv:2406.02631},
  year={2024}
}

@article{wang2024havtr,
  title={HaVTR: Improving Video-Text Retrieval Through Augmentation Using Large Foundation Models},
  author={Wang, Yimu and Yuan, Shuai and Jian, Xiangru and Pang, Wei and Wang, Mushi and Yu, Ning},
  journal={arXiv preprint arXiv:2404.05083},
  year={2024}
}

@article{song2024animation,
  title={An Animation-based Augmentation Approach for Action Recognition from Discontinuous Video},
  author={Song, Xingyu and Li, Zhan and Chen, Shi and Cai, Xin-Qiang and Demachi, Kazuyuki},
  journal={arXiv preprint arXiv:2404.06741},
  year={2024}
}

@article{rosa2024video,
  title={Video Enriched Retrieval Augmented Generation Using Aligned Video Captions},
  author={Rosa, Kevin Dela},
  journal={arXiv preprint arXiv:2405.17706},
  year={2024}
}

@article{pu2023spatial,
  title={Spatial-temporal knowledge-embedded transformer for video scene graph generation},
  author={Pu, Tao and Chen, Tianshui and Wu, Hefeng and Lu, Yongyi and Lin, Liang},
  journal={IEEE Transactions on Image Processing},
  year={2023},
  publisher={IEEE}
}

@inproceedings{rodin2024action,
  title={Action Scene Graphs for Long-Form Understanding of Egocentric Videos},
  author={Rodin, Ivan and Furnari, Antonino and Min, Kyle and Tripathi, Subarna and Farinella, Giovanni Maria},
  booktitle={Proceedings of the IEEE/CVF Conference on Computer Vision and Pattern Recognition},
  pages={18622--18632},
  year={2024}
}

@article{chen2024curriculum,
  title={Curriculum Negative Mining For Temporal Networks},
  author={Chen, Ziyue and Zheng, Tongya and Song, Mingli},
  journal={arXiv preprint arXiv:2407.17070},
  year={2024}
}

@article{nguyen2024video,
  title={Video-Language Understanding: A Survey from Model Architecture, Model Training, and Data Perspectives},
  author={Nguyen, Thong and Bin, Yi and Xiao, Junbin and Qu, Leigang and Li, Yicong and Wu, Jay Zhangjie and Nguyen, Cong-Duy and Ng, See-Kiong and Tuan, Luu Anh},
  journal={arXiv preprint arXiv:2406.05615},
  year={2024}
}

@article{nguyen2024encoding,
  title={Encoding and Controlling Global Semantics for Long-form Video Question Answering},
  author={Nguyen, Thong Thanh and Hu, Zhiyuan and Wu, Xiaobao and Nguyen, Cong-Duy T and Ng, See-Kiong and Luu, Anh Tuan},
  journal={arXiv preprint arXiv:2405.19723},
  year={2024}
}

@article{nguyen2023demaformer,
  title={Demaformer: Damped exponential moving average transformer with energy-based modeling for temporal language grounding},
  author={Nguyen, Thong and Wu, Xiaobao and Dong, Xinshuai and Nguyen, Cong-Duy and Ng, See-Kiong and Tuan, Luu Anh},
  journal={arXiv preprint arXiv:2312.02549},
  year={2023}
}

@inproceedings{nguyen2024read,
  title={READ-PVLA: Recurrent Adapter with Partial Video-Language Alignment for Parameter-Efficient Transfer Learning in Low-Resource Video-Language Modeling},
  author={Nguyen, Thong and Wu, Xiaobao and Dong, Xinshuai and Le, Khoi M and Hu, Zhiyuan and Nguyen, Cong-Duy and Ng, See-Kiong and Luu, Anh Tuan},
  booktitle={Proceedings of the AAAI Conference on Artificial Intelligence},
  volume={38},
  number={17},
  pages={18824--18832},
  year={2024}
}

@article{nguyen2021contrastive,
  title={Contrastive learning for neural topic model},
  author={Nguyen, Thong and Luu, Anh Tuan},
  journal={Advances in neural information processing systems},
  volume={34},
  pages={11974--11986},
  year={2021}
}

@article{nguyen2022adaptive,
  title={Adaptive contrastive learning on multimodal transformer for review helpfulness predictions},
  author={Nguyen, Thong and Wu, Xiaobao and Luu, Anh-Tuan and Nguyen, Cong-Duy and Hai, Zhen and Bing, Lidong},
  journal={arXiv preprint arXiv:2211.03524},
  year={2022}
}

@article{nguyen2024topic,
  title={Topic Modeling as Multi-Objective Contrastive Optimization},
  author={Nguyen, Thong and Wu, Xiaobao and Dong, Xinshuai and Nguyen, Cong-Duy T and Ng, See-Kiong and Luu, Anh Tuan},
  journal={arXiv preprint arXiv:2402.07577},
  year={2024}
}

@article{nguyen2023improving,
  title={Improving multimodal sentiment analysis: Supervised angular margin-based contrastive learning for enhanced fusion representation},
  author={Nguyen, Cong-Duy and Nguyen, Thong and Vu, Duc Anh and Tuan, Luu Anh},
  journal={arXiv preprint arXiv:2312.02227},
  year={2023}
}

@inproceedings{wu2023infoctm,
  title     = {Infoctm: A mutual information maximization perspective of cross-lingual topic modeling},
  author    = {Wu, Xiaobao and Dong, Xinshuai and Nguyen, Thong and Liu, Chaoqun and Pan, Liang-Ming and Luu, Anh Tuan},
  year      = 2023,
  booktitle = {Proceedings of the AAAI Conference on Artificial Intelligence},
  volume    = 37,
  pages     = {13763--13771},
  url       = {https://arxiv.org/abs/2304.03544}
}

@inproceedings{wu2024modeling,
    title = "Modeling Dynamic Topics in Chain-Free Fashion by Evolution-Tracking Contrastive Learning and Unassociated Word Exclusion",
    author = "Wu, Xiaobao  and Dong, Xinshuai  and Pan, Liangming  and Nguyen, Thong  and Luu, Anh Tuan",
    editor = "Ku, Lun-Wei  and Martins, Andre  and Srikumar, Vivek",
    booktitle = "Findings of the Association for Computational Linguistics ACL 2024",
    month = aug,
    year = "2024",
    address = "Bangkok, Thailand and virtual meeting",
    publisher = "Association for Computational Linguistics",
    url = "https://aclanthology.org/2024.findings-acl.183",
    pages = "3088--3105"
}

@article{nguyen2024kdmcse,
  title={Kdmcse: Knowledge distillation multimodal sentence embeddings with adaptive angular margin contrastive learning},
  author={Nguyen, Cong-Duy and Nguyen, Thong and Wu, Xiaobao and Luu, Anh Tuan},
  journal={arXiv preprint arXiv:2403.17486},
  year={2024}
}
\newpage
\onecolumn

\end{document}